\documentclass[11pt]{article}
\usepackage{acl2016}
\usepackage{times}
\usepackage{url}
\usepackage{latexsym}
\usepackage{amsmath}
\usepackage{amssymb}
\usepackage{graphicx}
\usepackage{multicol}
\usepackage{amssymb}
\usepackage{url}
\usepackage[draft]{hyperref}
\usepackage{multirow}
\usepackage{adjustbox}
\usepackage{xcolor}

\aclfinalcopy
\title{Guided Alignment Training for Topic-Aware Neural Machine Translation}

\author{
  \hspace*{7cm} Wenhu Chen$^2$, Evgeny Matusov$^1$, Shahram Khadivi$^1$, and Jan-Thorsten Peter$^2$ \\
  $^1$eBay, Inc. \\
  Kasernenstr. 25 \\
  52064 Aachen, Germany\\
  {\tt ematusov@ebay.com}, \\ 
  {\tt skhadivi@ebay.com} \And
  \ \\
  $^2$RWTH Aachen University \\
  Ahornstr. 55 \\
  52056 Aachen, Germany \\
  {\tt hustchenwenhu@gmail.com}, \\
  {\tt peter@cs.rwth-aachen.de}  \\
}
\date{}

\begin{document}
\maketitle
\begin{abstract}
In this paper, we propose an effective way for biasing the attention mechanism of a sequence-to-sequence neural machine translation (NMT) model towards the well-studied statistical word alignment models. We show that our novel guided alignment training approach improves translation quality on real-life e-commerce texts consisting of product titles and descriptions, overcoming the problems posed by many unknown words and a large type/token ratio. We also show that meta-data associated with input texts such as topic or category information can significantly improve translation quality when used as an additional signal to the decoder part of the network. With both novel features, the BLEU score of the NMT system on a product title set improves from 18.6 to 21.3\%. Even larger MT quality gains are obtained through domain adaptation of a general domain NMT system to e-commerce data. The developed NMT system also performs well on the IWSLT speech translation task, where an ensemble of four variant systems outperforms the phrase-based baseline by 2.1\% BLEU absolute. 

\end{abstract}
\section{Introduction}
NMT systems were shown to reach state-of-the-art translation quality on tasks established in MT research community such as IWSLT speech translation task~\cite{cettoloEtAl:EAMT2012}. In this paper, we also apply NMT approach to e-commerce data: user-generated product titles and descriptions for items put on sale. 
Such data are very different from newswire and other texts typically considered in the MT research community. Titles in particular are short (usually fewer than 15 words), contain many brand names which often do not have to be translated, but also product feature values and specific abbreviations and jargon. Also, the vocabulary size is very large due to the large variety of product types, and many words are observed in the training data only once. At the same time, these data are provided with additional meta-information about the item (e.g.\, product category such as clothing or electronics), which can be used as context to perform topic/domain adaptation for improved translation quality.

At first glance, established phrase-based statistical MT approaches are well-suited for e-commerce data translation. In a phrase-based approach, singleton, but unambiguous words and phrases are usually translated correctly. Also, since the alignment between source and target words is available, it is possible to transfer certain entities from the source sentence to the generated target sentence ``in-context'' without translating them. Such entities can include numbers, product specifications such as ``5S'' or ``.35XX'', or brand names such as ``Samsung'' or ``Lenovo''. In training, these entities can be replaced with placeholders to reduce the vocabulary size.  

However, NMT approaches are more powerful at capturing context beyond phrase boundaries and were shown to better exploit available training data. They also successfully adapt themselves to a domain, for which only a limited amount of parallel training data is available~\cite{luong2015iwslt}. Also, previous research~\cite{mathur2015topic} has shown that it is difficult to obtain translation quality improvements with topic adaptation in phrase-based SMT because of data sparseness and a large number of topics (e.\,g.~corresponding to product categories), which may or may not be relevant for disambiguating between alternative translations or solving other known MT problems. In contrast, we expected NMT to better solve the topic adaptation problem by using the additional meta-information as an extra signal in the neural network. To the best of our knowledge, this is the first work where the additional information about the text topic is embedded into the vector space and used to directly influence NMT decisions. 

In an NMT system, the attention mechanism introduced in \cite{luong2014addressing} is important both for decoding as well as for restoration of placeholder content and insertion of unknown words in the right positions in the target sentence. To improve the estimation of the soft alignment, we propose to use the Viterbi alignments of the IBM model 4~\cite{Brown93themathematics} as an additional source of knowledge during NMT training. The additional alignment information helps the current system to bias the attention mechanism towards the Viterbi alignment. 


This paper is structured as follows. After an overview of related NMT work in Section~\ref{sec:relwork}, we propose a novel approach in Section~\ref{sec:alignment} on how to improve the NMT translation quality by combining two worlds: the phrase-based SMT and its statistical word alignment and the neural MT attention mechanism. In Section~\ref{sec:category}, we describe in more detail how topic information can benefit NMT. Section~\ref{sec:bootstrap} and Section~\ref{sec:domain} describes our domain adaptation approach. Experimental results are presented in Section~\ref{sec:experiments}. The paper is concluded with a discussion and outlook in Section~\ref{sec:conclusions}.

\section{Related Work}\label{sec:relwork}
Neural machine translation is mainly based on using recurrent neural networks to grasp long term dependencies in natural language. An NMT system is trained on end-to-end basis to maximize the conditional probability of a correct translation given a source sentence \cite{sutskever2014sequence}, \cite{bahdanau2014neural}, \cite{cho2014learning}. When using attention mechanism, large vocabularies \cite{DBLP:journals/corr/JeanCMB14}, and some other techniques, NMT is reported to achieve comparable translation quality to state-of-art phrase-based translation systems. Most NMT approaches are based on the encoder-decoder architecture \cite{cho2014properties}, in which the input sentence is first encoded into a fixed-length representation, from which the recurrent neural network decoder generates the sequence of target words. Since fixed-length representation cannot give enough information for decoding, a more sophisticated approach using attention mechanism is proposed by \cite{bahdanau2014neural}. In this approach, the neural network learns to attend to different parts of source sentence to improve translation quality. Since the source and target language vocabularies for a neural network have to be limited, the rare words problem deteriorates translation quality significantly. The rare word replacement technique using soft alignment proposed by \cite{luong2014addressing} gives a promising solution for the problem. Both encoder-decoder architecture and insertion of unknown words into NMT output highly rely on the quality of the attention mechanism, thus it becomes the crucial part of NMT. Some research has been done to refine it by \cite{luong2015effective}, who proposed global and local attention-based models, and \cite{trevor2016incorporating}, who used biases, fertility and symmetric bilingual structure to improve the attention model mechanism.

Research on topic adaptation most closely related to our work was performed by~\cite{hasler2014dynamic}, but the features proposed there were added to the log-linear model of a phrase-based system. Here, we use the topic information as part of the input to the NMT system. Another difference is that we primarily work with human-labeled topics, whereas in~\cite{hasler2014dynamic} the topic distribution is inferred automatically from data. 

When translating e-commerce content, we are faced with a situation when only a few product titles and descriptions were manually translated, resulting in a small in-domain parallel corpus, but a large general-domain parallel corpus is available. In such situations, domain adaption techniques have been used both in phrase-based systems \cite{koehn2007experiments} and NMT \cite{luong2015iwslt}. In addition, while diverse NMT models using different features and techniques are trained, an ensemble decoder can be used to combine them together to make a more robust model. This approach was used by \cite{luong2015effective} to outperform the state-of-art phrase-based system with their NMT approach in the WMT 2015 evaluation.

\section{Guided Alignment Training}\label{sec:alignment}

When using the attention-based NMT \cite{bahdanau2014neural}, we observed that the attention mechanism sometimes fails to yield appropriate soft alignments, especially with increasing length of the input sentence and many out-of-vocabulary words or placeholders. In translation, this can lead to disordered output and word repetition.   

In contrast to a statistical phrase-based system, the NMT decoder does not have explicit information about the candidates of the current word, so at each recurrent step, the attention weights only rely on the previously generated word and decoder/encoder state, as depicted in \autoref{fig: category-aware-model}. The target word itself is not used to compute its attention weights. If the previous word is an out-of-vocabulary (OOV) or a placeholder, then the information it provides for calculating the attention weights for the current word is neither sufficient nor reliable anymore. This leads to incorrect target word prediction, and the error propagates to the future steps due to feedback loop. The problem is even larger in the case of e-commerce data where the number of OOVs and placeholders is considerably higher. 

To improve the estimation of the soft alignment, we propose to use the Viterbi alignments of the IBM model 4 as an additional source of knowledge during the NMT training. Therefore, we firstly extract Viterbi alignments as trained by the GIZA++ toolkit~\cite{och03:asc}, then use them to bias the attention mechanism. Our approach is to optimize on both the decoder cost and the divergence between the attention weights and the alignment connections generated by statistical alignments. The multi-objective optimization task is then expressed as a single-objective one by means of linear combination of two loss functions: the original and the new alignment-guided loss.

\subsection{Decoder Cost}
NMT proposed by \cite{bahdanau2014neural} maximizes the conditional log-likelihood of target sentence $y_1,\dots,y_T$ given the source sentence $x_1,\dots,x_T'$:
\begin{align}
	H_D(y,x) = -\frac{1}{N} \sum_{n=1}^N \log p_{\theta}(y_n|x_n)
    \label{eq: maximum-likelihood}
\end{align}
where $(y_n, x_n)$ refers to $n_{th}$ training sentence pair, and $N$ denotes the total number of sentence pairs in the training corpus. In the paper, we name the negative log-likelihood as decoder cost to distinguish from alignment cost. When using encoder-decoder architecture by \cite{cho2014learning}, the conditional probability can be written as: 
\begin{align}
\begin{split}
	p(y_1 \dots y_T | x_1 \dots x_{T'}) &= \prod_{t=1}^{T} p(y_t|y_{t-1} \cdots y_1, c)  
\end{split}
\label{eq: enc-dec-prob}
\end{align}
with 
\begin{align}
p(y_t|y_{t-1} \cdots y_1, c) = g(s_{t}, y_{t-1}, c) \nonumber
\end{align}
where $T$ is the length of the target sentence and $T'$ is the length of source sentence, $c$ is a fixed-length vector to encode source sentence, $s_{t}$ is a hidden state of RNN at time step $t$, and $g(\cdot)$ is a non-linear function to approximate word probability. If attention mechanism is used, the fixed-length $c$ is replaced by variable-length representation $c_t$ that is a weighted summary over a sequence of annotations $(h_1,\cdots,h_{T'})$, and $h_i$ contains information about the whole input sentence, but with a strong focus on the parts surrounding the $i_{th}$ word~\cite{bahdanau2014neural}. Then, the context vector can be defined as:
\begin{align}
c_t &= \sum_{i}^{T'} \alpha_{ti} h_i
\label{eq:ct}
\end{align}
where $\alpha_{ti}$ for each annotation $h_i$ is computed by normalizing the score function with softmax, as described in \autoref{eq: attention weights}. 
\begin{align}
    \alpha_{ti} = \frac{exp(e_{ti})}{\sum_{k=1}^{T'} exp(e_{tk})}
    \label{eq: attention weights}
\end{align}
Here $e_{ti} = a(s_{t-1}, h_i)$ is the function to calculate the score of $t$-th target word aligning to $i$-th word in the source sentence. The alignment model $a(\cdot,\cdot)$ is used to calculate similarity between previous state $s_{t-1}$ and bi-directional state $h_i$. In our experiments, we took the idea of dot global attention model \cite{luong2015effective}, but we still keep the order $h_{t-1} \rightarrow a_t \rightarrow c_t \rightarrow h_t$ as proposed by \cite{bahdanau2014neural}. We calculate the dot product of encoder state $h_i$ with the last decoder state $s_{t-1}$ instead of the current decoder state. We observe that this dot attention model (\autoref{eq: dot product}) works better than concatenation in our experiments.
\begin{align}
	a(s_{t-1}, h_i) = (W_s s_{t-1})^T (W_h h_i)
    \label{eq: dot product}
\end{align}

\subsection{Alignment Cost}\label{sec:alcost}

We introduce alignment cost to penalize attention mechanism when it is not consistent with statistical word alignment. We represent the pre-trained statistical alignments by a matrix $A$, where $A_{ti}$ refers to the probability of the $t_{th}$ word in the target sentence of being aligned to the $i_{th}$ word in the source sentence. In case of multiple source words aligning to the same target word, we normalize to make sure $\sum_i A_{ti} = 1$. In attention-based NMT, the matrix of attention weights $\alpha$ has the same shape and semantics as $A$. We propose to penalize NMT based on the divergence of the two matrices during the training, the divergence function can e.\,g.~be cross entropy $G_{ce}$ or mean square error $G_{mse}$ as in \autoref{eq: alignment-cost}. As shown in \autoref{fig: category-aware-model}, $A$ comes from statistical alignment, feeding into our guided-alignment NMT as an additional input to penalize the attention mechanism.
\begin{align}
\begin{split}
    G_{ce}(A, \alpha) &=  -\frac{1}{T}\sum_{t=1}^T \sum_{i=1}^{T'} A_{ti} \log \alpha_{ti} \\
    G_{mse}(A, \alpha) &= \frac{1}{T}\sum_{t=1}^T \sum_{i=1}^{T'}(A_{ti} - \alpha_{ti})^2 
\end{split}    
    \label{eq: alignment-cost}
\end{align}
We combine decoder cost and alignment cost to build the new loss function $H(y, x, A, \alpha)$:
\begin{align}
    H(y, x, A, \alpha) = w_1 G(A, \alpha) + w_2 H_D(y, x)
    \label{eq: combine-alignment}
\end{align}
 During training, we optimize the new compound loss function $H(y, x, A, \alpha)$ with regard to the same parameters as before. The guided-alignment training influences the attention mechanism to generate alignment closer to Viterbi alignment and has the advantage of unchanged parameter space and model complexity. When training is done, we assume that NMT can generate robust alignment by itself, so there is no need to feed an alignment matrix as input during evaluation. As indicated in \autoref{eq: combine-alignment}, we set $w_1$ and $w_2$ for weights of decoder cost and alignment cost to balance their weight ratio. We performed further experiments (see \autoref{sec:experiments}) to analyze the impact of different weight settings on translation quality.

\section{Topic-aware Machine Translation}\label{sec:category}
In the e-commerce domain, the information on the product category (e.g., ``men's clothing'', ``mobile phones``, ``kitchen appliances'') often accompanies the product title and description and can be used as an additional source of information both in the training of a MT system and during translation. In particular, such meta-information can help to disambiguate between alternative translations of the same word that have different meaning. The choice of the right translation often depends on the category. For example, the word ``skin'' has to be translated differently in the categories ``mobile phone accessories'' and ``make-up''. Outside of the e-commerce world, similar topic information is available in the form of e.g. tags and keywords for a given document (on-line article, blog post, patent, etc.) and can also be used for word sense disambiguation and topic adaptation. In general, the same document can belong to multiple topics. 

Here, we propose to feed such meta-information into the recurrent neural network to help generate words which are appropriate given a particular category or topic. 

\subsection{Topic Representation}
The idea is to represent topic information in a $D$-dimensional vector $l$, where $D$ is the number of topics. Since one sentence can belong to multiple topics (possibly with different probabilities/weights), we normalize the topic vector so that the sum of its elements is 1. It is fed into the decoder to influence the proposed target word distribution. The conditional probability given the topic membership vector can be written as (cf.~Equations~\ref{eq: enc-dec-prob} and~\ref{eq:ct}):
\begin{align}
\begin{split}
	p(y_t| y_{<t-1}, c_t, s_{t-1}, l) &= p(y_t | y_{t-1}, c_t, s_{t-1}, l)\\
									  &= g(y_{t-1}, s_{t-1}, c_t, l)
\end{split}
\nonumber
\end{align}
where $g(\cdot)$ is used to approximate the probability distribution. In our implementation, we introduce an intermediate readout layer to build function $g(\cdot)$, which is a feed-forward network as depicted in \autoref{fig: readout-layer}.

\subsection{Topic-aware Decoder}
	
In the NMT decoder, we feed the topic membership vector to the readout layer in each recurrent step to enhance word selection. As shown in \autoref{fig: category-aware-model}, topic membership vector $l$ is fed into NMT decoder as an additional input besides source and target sentences: 

\begin{figure*}[tb]
\includegraphics[width=0.60\linewidth]{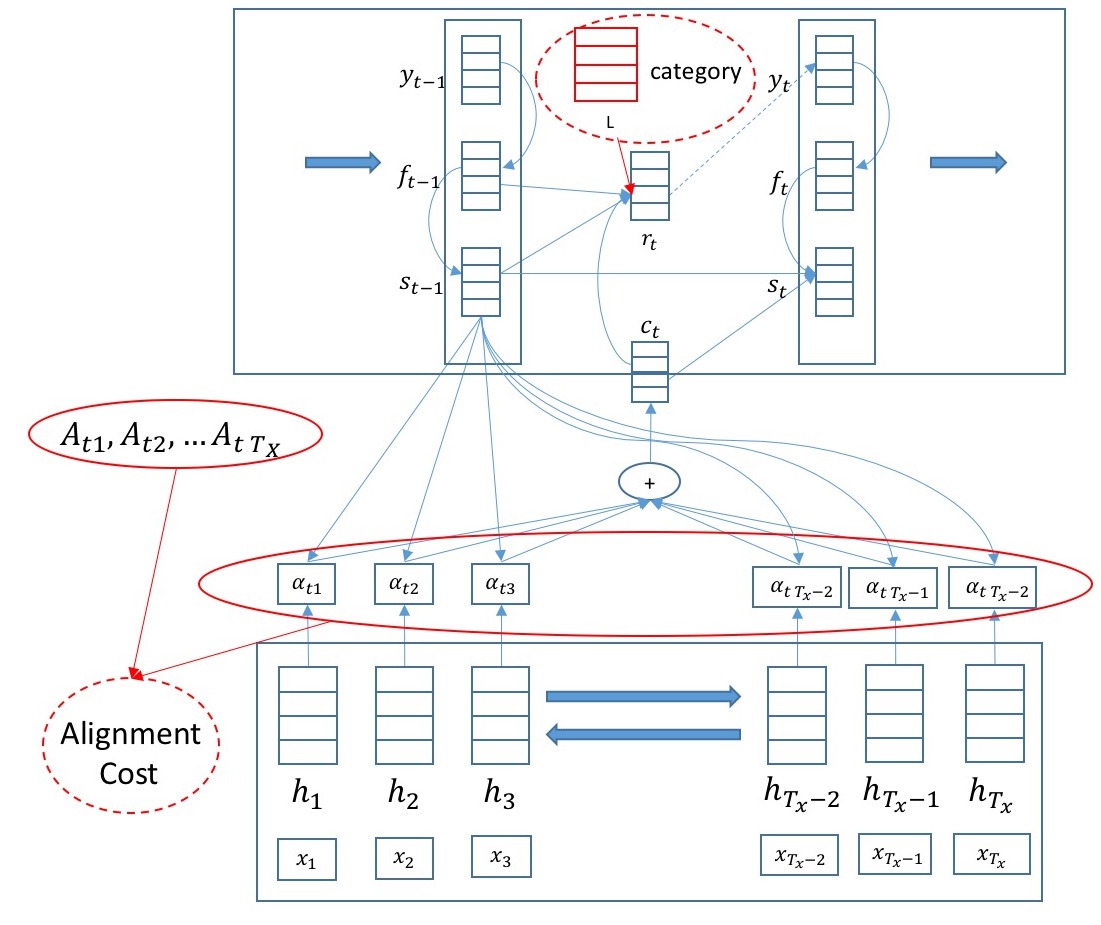}
\centering
\caption{Topic-aware, alignment-guided encoder-decoder model.}
\label{fig: category-aware-model}
\end{figure*}
\begin{figure}[ht]
\includegraphics[width=0.90\linewidth]{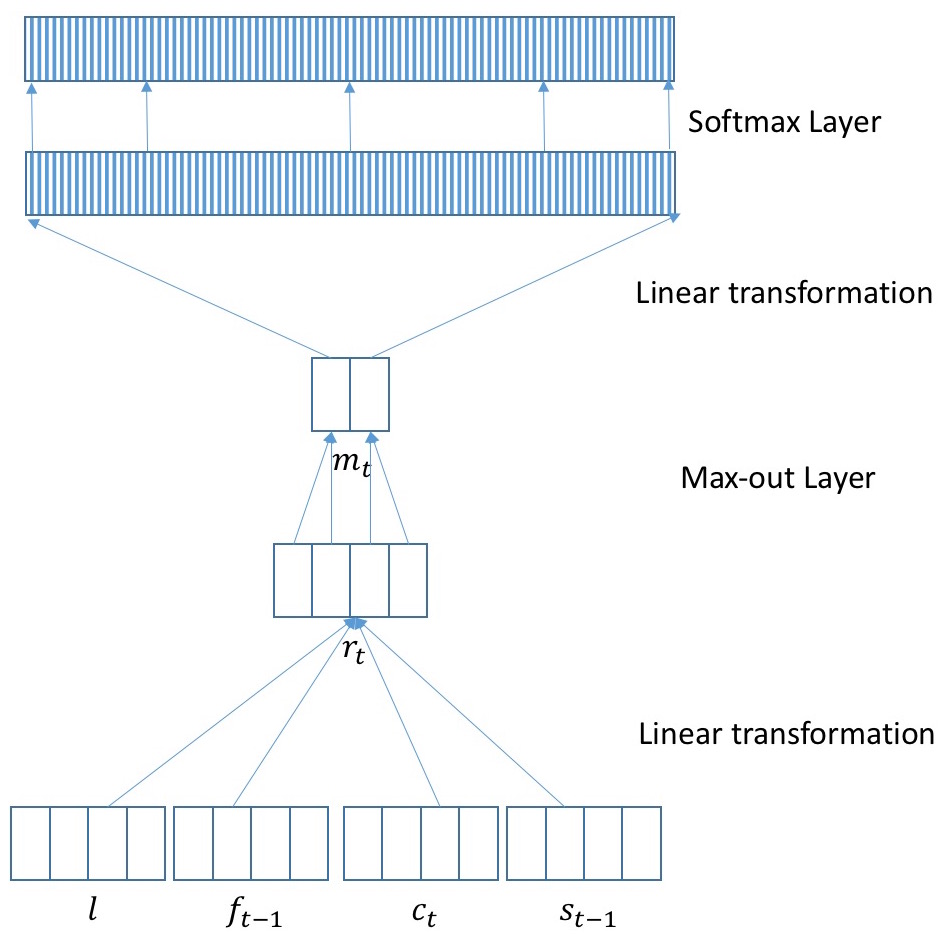}
\centering
\caption{Topic-aware readout layer.}
\label{fig: readout-layer}
\end{figure}

\begin{align}
p(y_t|y_{<t-1}, c_t, s_{t-1}, l) &= p(y_t|r_t) 
\label{eq: concatenate}
\end{align}
\begin{align}
r_t &= W_r [c_t;f_{t-1};s_{t-1};l] + b_r
\end{align}
where $W_r$ is concatenation of original transformation matrix and $l$, $r_t$ is the output from readout layer and $f_t$ is the embedding of the last target word $y_{t-1}$; $s_{t-1}$ refers the last decoder state. $W_r$ and $b_r$ are weights and bias for the linear transformation, respectively.
We can rearrange the formula as:
\begin{align}
\begin{split}
r_t &= [W_r', W_c][c_t;f_{t-1};s_{t-1};l] + b_r \\
    &= W_r' [c_t;f_{t-1};s_{t-1}] + W_c l + b_r \\
    &= W_r' [c_t;f_{t-1};s_{t-1}] + E_c + b_r \\
    &= r'_t + E_c
\end{split}
\label{eq: rearrange}
\end{align}
where $W_r$ is concatenation of original transformation matrix $W_r'$ and topic transformation matrix $W_c$. Then adding topic into readout layer input is equivalent to adding an additional topic vector $E_c$ into the original readout layer output. Assuming $l$ is a one-hot category vector, then $W_c l$ is equivalent to retrieving a specific column from the matrix $W_c$. Hence, we can name this additional vector $E_c$ as topic embedding, regarded as a vector representation of topic information. It is quite similar to word embedding by \cite{mikolov2013efficient}, we will further analyze the similarity between different topics in \autoref{fig: categ-correlation}.

The readout layer depicted in \autoref{fig: readout-layer} merges information from the last state $s_{t-1}$, previous word embedding $f_{t-1}$ (coming from word index $y_{t-1}$, which is sampled w.r.t.~the proposed word distribution), as well as the current context $c_t$ to generate output. It can be seen as a shallow network, which consists of a max-out layer \cite{goodfellow2013maxout}, a fully-connected layer, and a softmax layer.

\section{Bootstrapping}\label{sec:bootstrap}

When trained on small amounts of data, the attention-based neural network approach does not always produce reliable soft alignment. The problem gets worse when the sentence pairs available for training are getting longer. To solve this problem, we extracted bilingual sub-sentence units from existing sentence pairs to be used as additional training data. These units are exclusively aligned to each other, i.\,e.~all words within the source sub-sentence are aligned only to the words within the corresponding target sub-sentence and vice versa. The alignment is determined with the standard approach (IBM Model 4 alignment trained with the GIZA++ toolkit \cite{och03:asc}). As boundaries for sub-sentence units, we used punctuation marks, including period, comma, semicolon, colon, dash, etc. To simplify bilingual sentence splitting, we used the standard phrase pair extraction algorithm for phrase-based SMT, but set the minimum/maximum source phrase length to 8 and 30 tokens, respectively. From all such long phrase pairs extracted by the algorithm, we only kept those which are started or ended with a punctuation mark or started/ended a sentence; both on the source and on the target side. 

For the bootstrapped training, we merged the original training data with the extracted sub-sentence units and ran the neural training algorithm on this extended training set. Since the extracted bilingual sub-sentence units generally showed good correspondence between source and target due to the constraints described above, the expectation was that having such units repeated in the training data as stand-alone training instances would guide the attention mechanism to become more robust and make it easier for the neural training algorithm to find better correspondences between more difficult source/target sentence parts. Also, having both short and long training instances was expected to make neural translation quality less dependent on the input length.

\section{E-commerce Domain Adaptation}\label{sec:domain}
For the e-commerce English-to-French translation task, we only had a limited amount of in-domain parallel training data (item titles and descriptions). To benefit from large amounts of general-domain training data, we followed the method described in~\cite{luong2015iwslt}. We first trained a baseline NMT model on English-French WMT data (common-crawl, Europarl v7, and news commentary corpora) for two epochs to get the best result on a development set, and then continued training the same model on the in-domain training set for a few more epochs. In contrast to~\cite{luong2015iwslt}, however, we used the vocabularies of the most frequent 52K source/target words in the in-domain data (instead of the out-of-domain data vocabularies). This allowed us to focus the NN on translation of the most relevant in-domain words.

\section{Experimental Results}\label{sec:experiments}

\subsection{Datasets and Preprocessing}
We performed MT experiments on the German-to-English IWLST 2015 speech translation task~\cite{cettoloEtAl:EAMT2012} and on an in-house English-to-French e-commerce translation task. As part of data preprocessing, we tokenized and lowercased the corpora, as well as replaced numbers, product specifications, and other special symbols with placeholders such as \$num. We only keep these placeholders in training, but preserve their content as XML markups in the dev/test sets, which we try to restore using attention mechanism. This content is inserted for the generated placeholders on the target side based on the attention mechanism (see \cite{luong2014addressing}). In the beam search for the best translation, we make sure that each placeholder content is used only once. Using the same mechanism, we also pass OOV words to the target side ``as is'' (without using any special unknown word symbol). 

On both tasks, we evaluate all systems and system variants using case-insensitive BLEU~\cite{papineni2002bleu} and TER~\cite{snover2006study} scores on held-out development and test data using a single human reference translation.


\begin{table*}[htb]
\centering
\begin{tabular}{|c|c|c|c|c|c|}
\hline
\multicolumn{2}{|c|}{Data-set}                     & \multicolumn{2}{c|}{IWSLT} & \multicolumn{2}{c|}{e-commerce} \\ \hline
\multicolumn{2}{|c|}{Language}                     & German       & English        & English          & French           \\ \hline
\multirow{3}{*}{Training} & Sentences              & \multicolumn{2}{|c|}{165\,201}  & \multicolumn{2}{|c|}{516\,000}           \\ \cline{2-6} 
                          & Running words  performed on         & 3\,873\,816       & 3\,656\,038       & 2\,592\,202          & 2\,895\,089          \\ \cline{2-6} 
                          & Full vocabulary        & 103\,390    &    45\,068     & 119\,607           & 129\,848           \\ \cline{2-6} 
\multirow{2}{*}{Dev}      & Sentences              & \multicolumn{2}{|c|}{567}    & \multicolumn{2}{|c|}{910}             \\ \cline{2-6} 
                          & Running words          & 9\,812        &  10\,695      & 10\,339            & 11\,283            \\ \cline{2-6} 
\multirow{3}{*}{Test}     & Sentences              & \multicolumn{2}{|c|}{1100}          & \multicolumn{2}{|c|}{910}            \\ \cline{2-6} 
                          & Running words          & 19\,019       &  22\,895        & 10\,817            & 11\,016            \\ 
\hline
\multicolumn{2}{|c|}{Source OOV rate \% w.r.t. full/NMT vocabulary} &   
\multicolumn{2}{|c|}{5.16/6.12}    &  \multicolumn{2}{|c|}{2.56/5.76} \\ \hline
\end{tabular}
\caption{Corpus statistics for the IWSLT and e-commerce translation tasks. OOV rate is calculated after preprocessing, placeholders like \$num, \$url, etc. largely decrease the OOV rate in the e-commerce dev and test sets.}
\label{tab: data-desc}
\end{table*}

\subsubsection{IWSLT TED Talk Data}
For the IWSLT German-to-English task (translation of transcribed TED talks), we mapped the topic keywords of each TED talk in the 2015 training/dev/test evaluation campaign release to ten general topics such as politics, environment, education, and others. All sentences in the same talk share the same topic, and one talk can belong to several topics. Instead of using the official IWSLT dev/test data, we set aside 81/159 talks for development/test set, respectively. Out of these talks, we used 567 dev and 1100 test sentences which had the highest probability of relating to a particular topic (bag-of-words classification using the remaining 1365 talks as the training data). The full corpus statistics for the IWSLT data sets obtained this way are given in \autoref{tab: data-desc}\footnote{This IWSLT data set with topic labels is publicly available at https://github.com/wenhuchen/iwslt-2015-de-en-topics.}. 

\subsubsection{E-commerce Data}
For the e-commerce English-to-French task, we used the product category such as ``fashion'' or ``electronics'' as topic information (a total of 80 most widely used categories plus the category ``other'' that combined all the less frequent categories). The training set contained both product titles and product descriptions, while dev and test set only contained product titles. Each title or description sentence was assigned to only one category. The statistics of the e-commerce data sets are given in \autoref{tab: data-desc}.
%

\subsection{Model Training}
We implemented our neural translation model in Python using the Blocks deep learning library \cite{van2015blocks} based on the open-source MILA translation project. We compared our implementation of NMT baseline system with \cite{bahdanau2014neural} on the WMT 2014 English-to-French machine translation task and obtained a similar BLEU score on the official test set as they reported in \cite{bahdanau2014neural}. Then we implemented the topic-aware algorithm  (\autoref{sec:category}), guided alignment training (\autoref{sec:alignment}), and the bootstrapped training (\autoref{sec:bootstrap}) into the NMT model. We trained separate models with various feature combinations.
We also created an ensemble of different models to obtain the best NMT translation results. 

In our experiments, we set the word embedding size to 620 and used a two-layer bi-directional GRU encoder and one layer of GRU decoder, the cell dimension of both were 1000. We selected the 50k most frequent German words and top 30k English words as vocabularies for the IWSLT  task, and most frequent 52k English/French words for the e-commerce task.  The optimization of the objective function was performed by using AdaDelta algorithm \cite{zeiler2012adadelta}. We set the beam size to 10 for dev/test set beam search translation.

For training implementation, we use stochastic gradient descent with batch size of 100, saving model parameters after a certain number of epochs. We saved around 30 consecutive model parameters. We selected the best parameter set according to the sum of the established MT evaluation measures BLEU~\cite{papineni2002bleu} and 1-TER~\cite{snover2006study} on the development set.
After model selection, we evaluated the best model on the test set. We report the test set BLEU and TER scores in \autoref{tab: final-result-ebay} and \autoref{tab: final-result-ted}. 

We use TITAN X GPUs with 12GB of RAM to run experiments on Ubuntu Linux 14.04. The training converges in less than 24 hours on the IWSLT talk task and around 30 hours on the e-commerce task. The beam search on the test set for both tasks takes around 10 minutes, the exact time depends on the vocabulary size and beam size.

\begin{table}[t]
\centering
\begin{tabular}{|l|c|c|}
\hline
E-commerce En$\rightarrow$Fr  & BLEU & TER  \\
							  & \% &  \%   \\ \hline
Baseline NMT       			  & 18.6  & 68.5  \\ \hline
+prefixed human-labeled categs & 18.3  & 69.3 \\ \hline 
+readout human-labeled categs  & 19.7  & 65.3  \\ \hline
+readout LDA topics 	    & 14.5 & 74.9 \\ \hline  
\end{tabular}
\caption{Comparison of different approaches for topic-aware NMT.}
\label{tab:categ-comparison}
\end{table}

\subsection{Effect of Topic-aware NMT}
We tested different approaches to find out where topic information fits best into NMT, since topic information can affect alignment, word selection, etc. The most naive approach is to insert a pseudo topic word in the beginning of a sentence to bias the context of the sentence to a certain topic. We also tried topic vectors of different origin in the read-out layer of the network. We used both topics predicted automatically with the Latent Dirichlet Analysis (LDA) and human-labeled topics to feed into the network as shown in \autoref{fig: category-aware-model}. 
\begin{table*}[ht]
\centering
\small
\begin{tabular}{|l|l|}
\hline
source & ich m\"ochte Ihnen heute Morgen gerne von meinem Projekt, Kunst Aufr\"aumen, erz\"ahlen.\\
NMT & I want to clean you this morning, from my project, to say Art. \\
+ topics & I would like to talk to you today by my project, Art clean. \\ 
reference & I would like to talk to you this morning about my project, Tidying Up Art. \\
\hline
\hline
source & \dots\ unsere Kollegen an Tufts verbinden Modelle wie diese mit durch Tissue Engineering erzeugten Knochen,  \\
 & um zu sehen, wie Krebs sich von einem Teil des K\"orpers zum n\"achsten verbreiten k\"onnte.\\
NMT & \dots\ our NOAA colleagues combined models of models like this with tissue generated bones
from bones to see \\
 & how cancer could spread from one part of the body, to the next distribution.\\
+ topics & \dots\ our colleagues at Tufts are using models like this with tissue-based engineered bones to see \\
 & how cancer could spread from a part of the body to the next part. \\
reference & \dots\ our colleagues at Tufts are mixing models like these with tissue-engineered bone to see \\
 & how cancer might spread from one part of the body to the next. \\
\hline
\end{tabular}
\caption{Example of improved translation quality when topic information is used as input in the neural MT system (German-to-English IWSLT test set).}
\label{tab:examples-category}
\normalsize
\end{table*}

\begin{figure}[tb]
\includegraphics[width=0.5\textwidth]{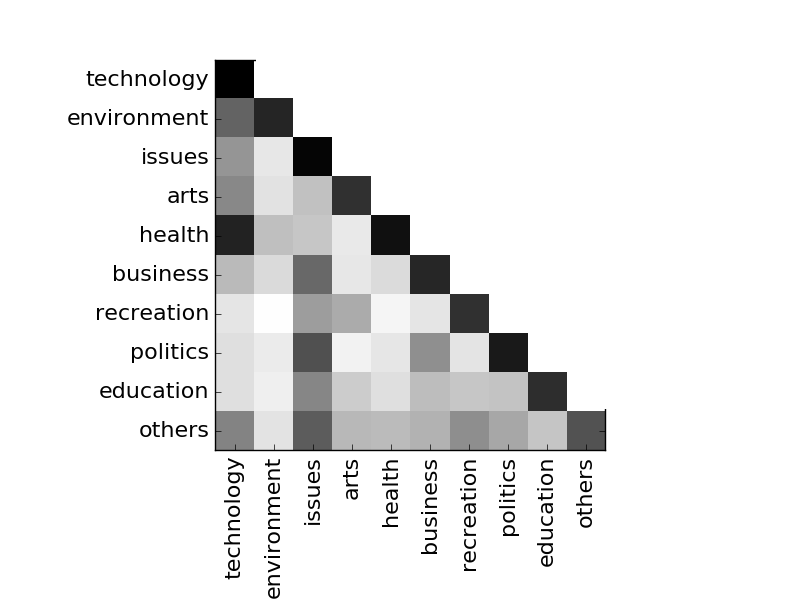}
\centering
\caption{Topic embedding cosine distance.}
\label{fig: categ-correlation}
\end{figure}

\begin{table}[tb]
\begin{tabular}{|l|l|l|}
\hline
E-commerce En$\rightarrow$Fr & BLEU \% & TER \%   \\ \hline
Baseline NMT          & 18.6  & 68.3 \\ \hline
+cross-entropy (decay)& 20.5  & 65.8 \\ \hline
+cross-entropy (1:2)  & 20.6  & 65.5 \\ \hline
+cross-entropy (1:1)  & 20.2  & 65.0 \\ \hline
+cross-entropy (2:1)  & 20.9  & 65.7 \\ \hline
+squared error (1:1)  & 20.8  & 64.5 \\ \hline
\end{tabular}
\caption{Comparison of different loss functions and weight ratios for guided alignment.}
\label{tab: alignment-comparison}
\end{table}

\begin{figure}[h]
\includegraphics[width=0.7\linewidth]{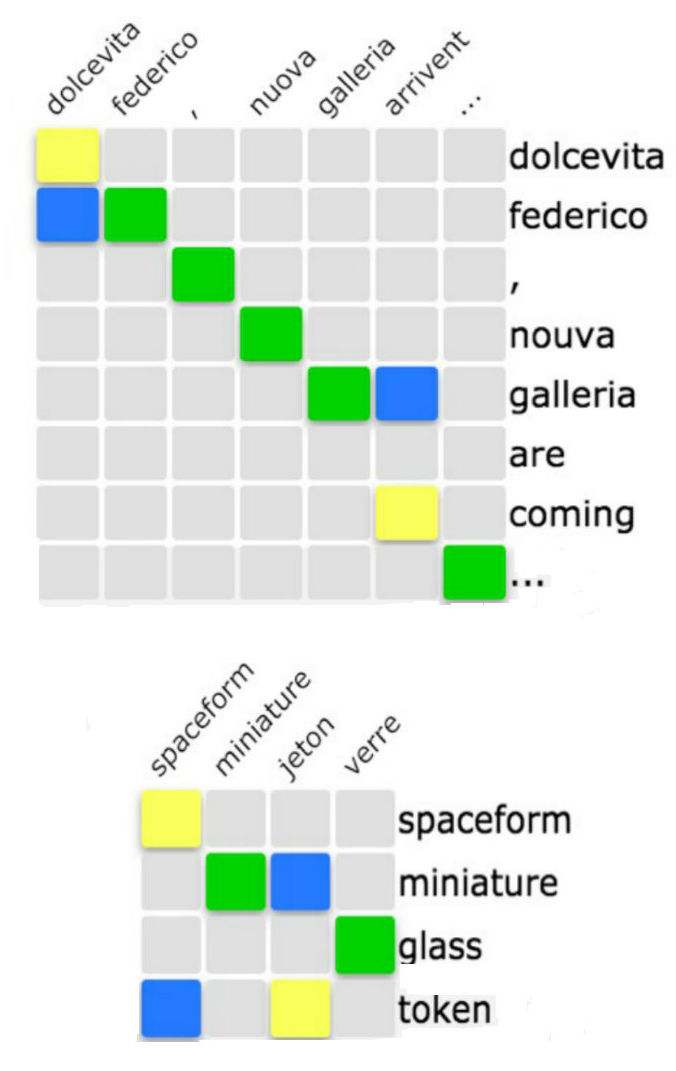}
\centering
\caption{Refined alignment examples using guided-alignment learning (green blocks refer to the identical alignments from Baseline NMT and guided-alignment NMT, blue blocks refer to the alignment from baseline NMT, yellow blocks refer to guided-alignment NMT).}
\label{fig: alignment-example}
\end{figure}

The results on the e-commerce task in Table~\ref{tab:categ-comparison} show that category information as a pseudo topic word does not carry enough semantic and syntactic meaning in comparison to real source words to have a positive effect on the target words predicted in the decoder. The BLEU score of such system (18.3\%) is even below the baseline (18.6\%). In contrast, the human-labeled categories are more reliable and are able to positively influence word selection in the NMT decoder, significantly (19.7\% BLEU) outperforming the baseline. 

Replacing the human-labeled topic one-hot vectors of size 80 with the LDA-predicted topic distribution vectors of the same dimension in the read-out layer of the neural network deteriorated the BLEU and TER scores significantly. We attribute this to data sparseness problems when training the LDA of dimension 80 on product titles.

On the German-to-English task, we also observed MT quality improvements when using human-labeled topic information as described in \autoref{fig: category-aware-model}. Here, we extracted the topic embedding $E_c$ from different experiments and show their cosine distance
in \autoref{fig: categ-correlation}. It's straightforward that in different experiments, the same topic tends to share similar representation in continuous embedding space. At the same time, closer topic pairs like ``politics'' and ``issues'' tend to have shorter distance from each other.  Examples of improved German-to-English NMT translations when human-labeled topic information is used are shown in Table~\ref{tab:examples-category}.

\begin{table}[t]
\centering
\begin{tabular}{|l|l|c|c|}
\hline
\multicolumn{2}{|l|}{\multirow{2}{*}{System Description} } & {\small BLEU}  & {\small TER}       \\ 
 \multicolumn{2}{|l|}{}  & \%   & \%       \\ \hline \hline
 \multicolumn{2}{|l|}{1. NMT in-domain (ID)}       & 18.6      & 68.5         \\ \hline
 \multicolumn{2}{|l|}{2. 1) + topic vectors} & 19.7     & 65.3         \\ \hline
 \multicolumn{2}{|l|}{3. 1) + bootstrapping} & 20.1   & 66.2  \\ \hline
\multicolumn{2}{|l|}{4. 1) + guided alignment} & 20.9 & 65.7  \\ \hline
\multicolumn{2}{|l|}{5. NMT with 2) and 4)} & 21.3 & 64.3 \\ \hline
\multicolumn{2}{|l|}{6. NMT with 2) and 3) and 4)} & 20.7  & 66.2 \\ \hline \hline
\multicolumn{2}{|l|}{7. NMT out-of-domain (OOD) }   & 13.8      & 77.4         \\ \hline
\multicolumn{2}{|l|}{8. 7) + guided alignment} & 16.3  & 74.5  \\ \hline
\multicolumn{2}{|l|}{9. 8) + domain adaptation} & 25.0   & 60.1   \\ \hline \hline
\multirow{2}{*}{Ensemble}  & system 4) & \multirow{4}{*}{24.5}  & \multirow{4}{*}{60.9} \\ 
\cline{2-2} & system 5)         				 &                        &                       \\ 
\cline{2-2} NMT  & system 6)        &                        &                       \\ 
\cline{2-2} ID & NMT w.~3) and 4)		 & 					      & \\ \hline
\multirow{1}{*}{Ensemble}   
& system 9)   											& \multirow{3}{*}{25.6} & \multirow{3}{*}{58.6} \\ 
\cline{2-2}  NMT &  9) with DW         &                        &                       \\ 
\cline{2-2}  OOD & 9) w.~topic vectors    			&                        &                       \\ \hline
\end{tabular}
\caption{Overview of the translation results on the English-to-French e-commerce translation task. (DW: decaying weight for the statistical alignment).}
\label{tab: final-result-ebay}
\end{table}

\begin{table*}[t]
\centering
\small
\begin{tabular}{|l|l|}
\hline
source & Vintage Ollech \& Wajs Early Bird Diver watch, Excellent! \\
SMT &  Vintage Ollech \& Wajs d\'ebut oiseau montre de plong\'ee, excellent! \\
NMT & Montre de plong\'ee vintage Ollech \& Wajs early bird, excellent! \\ 
reference & Montre de Plong\'ee Vintage Ollech \& Wajs Early Bird, Excellent ! \\
\hline
\hline
source & APT Holman Model 1 Audiophile Power Amplifer made in Cambridge Mass \\
SMT & APT Holman modèle 1 audiophile power fabricant d'ampli made in Cambridge Mass \\
NMT & L'amplificateur de puissance audiophile APT Holman mod\`ele 1 fabriqu\'ee \'a Cambridge Mass\\
reference & Amplificateur de puissance pour audiophile APT Holman mod\`ele 1 fabriqu\'e \'a Cambridge Massachussets \\
\hline
\end{tabular}
\caption{Example of improved translation quality of the NMT ensemble system vs. phrase-based baseline system (English-to-French title test set).}
\label{tab:examples-pbt-nmt}
\normalsize
\end{table*}

\begin{table*}[t]
\centering
\begin{tabular}{|l|l|l|c|c|}
\hline
Exp                & \multicolumn{2}{l|}{System Description}         & BLEU \%             & TER \%      \\ \hline
1                  & \multicolumn{2}{l|}{Phrase-based system}        & 24.7                  & 55.4          \\ \hline
2                  & \multicolumn{2}{l|}{Phrase-based system + OSM}  & 25.7                  & 55.1          \\ \hline
3                  & \multicolumn{2}{l|}{NMT}                        & 23.4                  & 60.1          \\ \hline
4                  & \multicolumn{2}{l|}{NMT + topic vectors}             & 23.7                  & 59.6          \\ \hline
5                  & \multicolumn{2}{l|}{NMT + bootstrapping}        & 24.1                  & 58.6          \\ \hline
6                  & \multicolumn{2}{l|}{NMT + guided alignment}     & 23.8                  & 60.8          \\ \hline
7 				   & \multicolumn{2}{l|}{NMT + topic vectors + bootstrapping} & 24.2				 & 59.4			 \\ \hline
8                  & \multicolumn{2}{l|}{NMT + topic vectors + bootstrapping + guided alignment} & 24.6  & 57.7   \\ \hline
\multirow{4}{*}{9} 
& \multirow{4}{*}{Ensemble} 
& NMT + topic vectors & \multirow{4}{*}{27.8}  & \multirow{4}{*}{55.4} \\ 
\cline{3-3}
&       
& NMT + topic vectors + guided alignment 
&
& \\ \cline{3-3}
&
& NMT + topic vectors + bootstrapping 
& 
& \\ \cline{3-3}
&
& NMT + topic v.~+ guided alignment + bootstrapping  
&
& \\ \hline
\end{tabular}
\caption{Overview of the translation results on the English-to-German IWSLT task.}
\label{tab: final-result-ted}
\end{table*}

\subsection{Implementation of Guided Alignment}
To balance decoder cost and the attention weight cost, we experimented with different weights for these costs. We analyzed the relation between weight ratio and the final result in \autoref{tab: alignment-comparison}. Besides fixing the cost ratios during training, we also apply a heuristic to adjust the ratio as the training is progressing. One approach is to set a high value for the alignment cost in the beginning, then decay the weight to 90\% after every epoch, finally eliminating the influence of the alignment after some time. This approach helps for the IWSLT task, but not on the e-commerce task. We assume that the alignment for the TED talk sentences seems to be easier for NMT to learn on its own than the alignment between product titles and their translations. We also analyzed the effect of using different loss functions for calculating alignment divergence (see Section~\ref{sec:alcost}). 
The difference between the squared error and cross-entropy is not so large as shown in 
\autoref{tab: alignment-comparison}. Since the cross-entropy function has a consistent form as decoder cost, we decided to use it in further experiments. We extracted the NMT attention weights and marked the connection with the highest score as hard alignment for each word. We drew the alignment in \autoref{fig: alignment-example} to compare baseline NMT and alignment-guided NMT. It can be seen from the graph that the guided alignment training truly improves the alignment correspondence.

\subsection{Overall Results}
The overall results on the e-commerce translation task and IWSLT task are shown in \autoref{tab: final-result-ebay} and \autoref{tab: final-result-ted}, respectively. We observed consistency on both tasks, in a sense that a feature that improves BLEU/TER results on one task is also beneficial for the other. 

For comparison, we trained phrase-based SMT models using the Moses toolkit~\cite{koehn2007moses} on both translation tasks. We used the standard Moses features, including a 4-gram LM trained on the target side of the bilingual data, word-level and phrase-level translation probabilities, as well as the distortion model with the maximum distortion of 6. Our stronger phrase-based baseline included additional 5 features of a 4-gram operation sequence model -- OSM~\cite{durrani2015operation}.

On the e-commerce task, which is more challenging due to a high number of OOV words and placeholders, 
we observed that NMT translation output had many errors related to incorrect attention weights. To improve the attention mechanism, we applied guided alignment and bootstrapping. Both boosted the translation performance. Adding topic information increased the BLEU score to 21.3\%. We selected the four best model parameters from various experiments to make an ensemble system, this improved the BLEU score to 24.5\%. For the following experiment, we had pre-trained a model on WMT15 parallel data with the guided alignment technique, and then continued training on the e-commerce data for several epocs as described in \autoref{sec:domain}, performing domain adaptation. This approach proved to be extremely helpful, giving an increase of over 3.0\% absolute in BLEU. Finally, we also applied ensemble methods on variants of the domain-adapted models to further increase the BLEU score to 25.6, which is 7.0 BLEU higher than the NMT baseline system, and only 0.6\% BLEU behind the BLEU score of 26.2\% for the state-of-the-art phrase-based baseline. Table~\ref{tab:examples-pbt-nmt} shows examples where the ensemble NMT system is better than the phrase-based system despite the slightly lower corpus-level BLEU score. In fact, a more detailed analysis of the sentence-level BLEU scores showed that the  NMT translation of 386 titles out of 910 was ranked higher than the SMT translation, the reverse was true for 460 titles. In particular, the word order of noun phrases was observed to be better in the NMT translations.

On the IWSLT task (\autoref{tab: final-result-ted}), the baseline NMT was not as far behind the phrase-based system as on the e-commerce task, and thus the obtained improvements were smaller than for product title translations. We observed that topic information is less helpful than bootstrapping and guided alignment learning. When we combined them, we reached the same BLEU score as the phrase-based system (see \autoref{tab: final-result-ted}). Finally, we combined four variant systems to create an ensemble, which resulted in the BLEU score of 27.8\%, surpassing the phrase-based translation with the OSM model by 2.1\% BLEU absolute.

\section{Conclusion}\label{sec:conclusions}
We have presented a novel guided alignment training for a NMT model that utilizes IBM model 4 Viterbi alignments to guide the attention mechanism. This approach was shown experimentally to bring consistent improvements of translation quality on e-commerce and spoken language translation tasks. Also on both tasks, the proposed novel way of utilizing topic meta-information was shown to improve BLEU and TER scores. We also showed improvements when using domain adaptation by continuing training of an out-of-domain NMT system on in-domain parallel data. In the future, we would like to investigate how to effectively make use of the abundant monolingual data with human-labeled product category information that we have available for the envisioned e-commerce application.

\bibliographystyle{acl2016}
\bibliography{reference}

\end{document}